\documentclass{article}

\usepackage{arxiv}

\usepackage[utf8]{inputenc} 
\usepackage[T1]{fontenc}    
\usepackage{hyperref}       
\usepackage{url}            
\usepackage{booktabs}       
\usepackage{amsfonts}       
\usepackage{nicefrac}       
\usepackage{microtype}      
\usepackage{lipsum}		
\usepackage{graphicx}
\usepackage{natbib}
\usepackage{doi}
\usepackage{subcaption}
\usepackage{algorithm}
\usepackage{algorithmic}
\usepackage{xcolor}

\title{Chitrakar: Robotic System for Drawing Jordan Curve of Facial Portrait}

\author{Aniruddha Singhal \\
	TCS Robotics Research Lab\\
	New Delhi, India \\
	\texttt{aniruddha.singhal@tcs.com} \\
	\And
	Ayush Kumar \\
	TCS Robotics Research Lab\\
	New Delhi, India \\
	\texttt{ayush.kumar@tcs.com} \\
	\And
	Shivam Thukral \\
	University of British Columbia\\
	Vancouver, Canada \\
	\texttt{tshivam@cs.ubc.ca} \\
	\And
	Deepak Raina \\
	Indian Institute of Technology\\
	Delhi, India \\
	\texttt{deepak.raina@mech.iitd.ac.in} \\
	\And
	Swagat Kumar \\
	Edge Hill University\\
	Lancashire, United Kingdom \\
	\texttt{swagat.kumar@edgehill.ac.uk} \\
}

\renewcommand{\headeright}{}
\renewcommand{\undertitle}{}

\hypersetup{
pdftitle={Chitrakar: Robotic System for Drawing Jordan Curve of Facial Portrait},
pdfsubject={q-bio.NC, q-bio.QM},
pdfauthor={Aniruddha Singhal, Ayush Kumar, Shivam Thukral, Deepak Raina, Swagat Kumar},
pdfkeywords={Robot art, Image processing, Jordan curve, Aesthetics and Human face portrait},
}

\begin{document}
\maketitle

\begin{abstract}
This paper presents a robotic system (\textit{Chitrakar}) which autonomously converts any image of a human face to a recognizable non-self-intersecting loop (Jordan Curve) and draws it on any planar surface. The image is processed using Mask R-CNN for instance segmentation, Laplacian of Gaussian (LoG) for feature enhancement and intensity based probabilistic stippling for the image to points conversion.
These points are treated as a destination for a travelling salesman and are connected with an optimal path which is calculated heuristically by minimizing the total distance to be travelled. 
This path is converted to a Jordan Curve in feasible time by removing intersections using a combination of image processing, 2-opt, and Bresenham's Algorithm. 
The robotic system generates $n$ instances of each image for human aesthetic judgement, out of which the most appealing instance is selected for the final drawing. The drawing is executed carefully by the robot's arm using trapezoidal velocity profiles for jerk-free and fast motion. The drawing, with a decent resolution, can be completed in less than 30 minutes which is impossible to do by hand. 
This work demonstrates the use of robotics to augment humans in executing difficult craft-work instead of replacing them altogether. 
\end{abstract}


\section{Introduction} 
		

	With the advent of Computers and Robotics, the technique of visual art is undergoing a revolutionary change. Computers are used to create art-works based on abstract mathematical ideas such as fractals, strange attractors and Travelling Salesman Problem art (TSP-art) \citep{gleick2011chaos, doi:10.1080/17513472.2011.634320}. 
		Modern computers with high processing power can effortlessly generate multiple iterations of art-work in seconds. This abundance of computing power is a new tool in the hands of the artist. It is not just a physical tool to create art but a computational tool which can be used to expand the horizon of the artist's imagination and can create soft copies of art-forms. Thus,  artist can execute what was deemed impossible before the advent of computers \citep{cornock1973creative}. Robotics can be used to augment the human capability of the craft by long, precise and dextrous movement. A digitally generated visual art-work can be physically implemented on a canvas using a robotic arm instead of a human hand so that hard copies of art-form can be generated accurately and efficiently.

		The involvement of human in the creation of art-work is essential and they cannot be replaced by machines. Computers can enhance human imagination and robotics can augment the human capability of its craftful execution \citep{bowie2018aesthetics}.
	The ability of computers to create new art is amazing and terrifying at the same time \citep{harari2016homo}. It can be argued that computers can create art-work which will be more pleasing to the viewers as compared to what a human has created because computers can exhaust large datasets of individual preferences and likings \citep{harari2016homo}. The entertainment industry has already started to explore the use of computers in content creation and generation \citep{liao_2019}. With more advancement in computers, it can be expected from machines to create incrementally better content \citep{tan2016ceci}. 
		\begin{figure}
			\includegraphics[width=\textwidth]{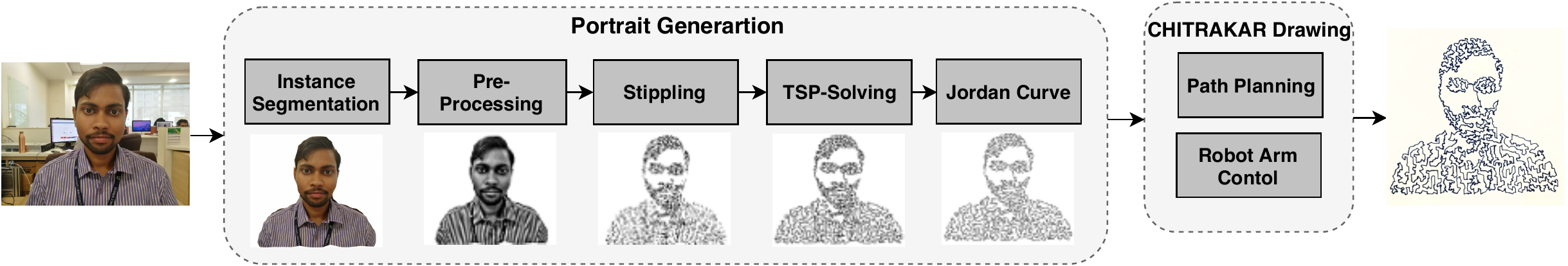}
			\caption{System Architecture: The input image taken by the camera mounted on top of robot is (1) segmented, (2) enhanced, (3) stippled, (4) connected through TSP-solver, (5) converted to multiple Jordan Curves and is sent for (6) trajectory planning to (7) control the robot arm for drawing the curve selected by human.}
			\label{fig:flow_chart}
		\end{figure}
		The other side of the story explains that the art is not separate from the observer itself and the observer invests his/her consciousness in interpreting the art which a machine cannot \citep{searle1980minds, nagel1974like}. Plato said that - ``Beauty is in the eyes of the beholder". To appreciate any form of art a qualification is required \citep{lothian1999landscape}. The qualification of the observer can be very abstract, like his/her ability to connect with the art-form he is looking at. 
			For example, in the case of TSP-art, an observer who does not appreciate the complexity of the problem of TSP and the beauty of Jordan Curve cannot see that a single non-intersecting loop is being used to make this type of art-work, and thus unable to appreciate its aesthetic value. A truly intelligent machine can create astonishing art-work but the appreciation and aesthetic judgement will always remain with the humans looking at it \citep{lothian1999landscape}.

\begin{figure}[h]
\centering
\includegraphics[width=0.6\linewidth, angle = 0]{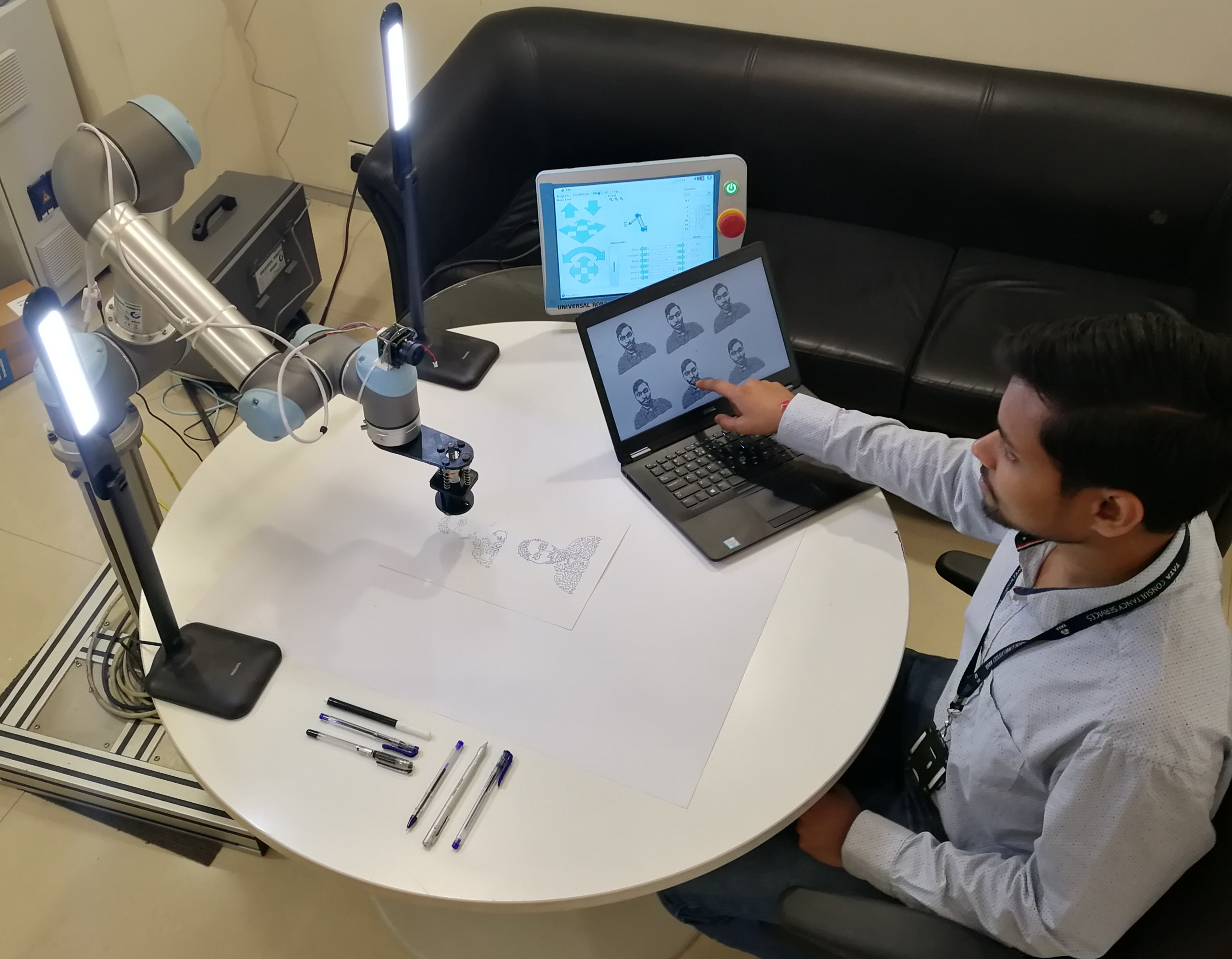}
\label{fig:real_system}
\caption{\small Human selecting the aesthetically appealing jordan curve to be drawn by the robotic system}
\label{fig:real_system}
\end{figure}


In this paper, we have developed an intelligent robotic system, called  \textit{Chitrakar} to convert any photograph of a human face into a portrait comprising of a single non-self-intersecting loop (Jordan Curve) and draw it on a plane surface (a paper on a table or a wall). The word ``Chitrakar'' in Hindi stands for ``Painter''. We are using this word to represent a robotic painter that can mimic the human ability to generate beautiful art. 
The robotic system takes a photograph of the subject (human face) and subtracts the background from the image using Mask R-CNN. In order to automate the complete pipeline, it is essential to know the pixel coordinates of the person in the image. This process of instance segmentation is done using Mask R-CNN. Remaining image processing pipeline can be applied to the segmented image without taking the background into consideration. The subtracted image is first transformed into a  grayscale image using Chromatic Adaptation Transform (CAT). The output is enhanced by applying the LoG transform and by adding the negative of the resultant image to the original. 
  The enhanced image is probabilistically stippled based on the intensity of pixels. The stippled image is converted to a set of the desired number of points which is fed to a TSP-solver. The solution of the TSP-solver is not always intersection free and, therefore it is sent for further processing for intersection detection and removal by using a novel approach proposed in this paper. The robotic system then repeats the process of stippling and generates $n$ different Jordan Curve out of which one is selected by a human observer based on his own sense of aesthetics. The selected curve is finally drawn by a UR5 robotic arm on the medium of choice which can be a paper or a wall.
  
We have demonstrated how computers, mathematics, and robotics can be combined to create a new form of art which was not possible before. 
Overall, the aim of this work is not to replace humans with robots but to assist them in difficult craft-work and use their subjectivity in the creation of visually appealing art. In short, the novel contributions made in this paper are: (1) We presented a novel approach to convert an image of a human face to a Jordan Curve using state-of-the-art deep-learning method and series of image processing techniques.
(2) We design a novel end-effector mechanism that allows a robotic arm to draw this portrait on a flat surface.  By using a robotic system we overcome the limitation of a human to precisely draw non-intersecting connections between thousands of point in feasible time. 
Thus, we demonstrate a method to integrate multi-disciplinary knowledge of robotics, mathematics, computer vision, art and aesthetics for development of a system to assist humans in the creation of unique artwork.

\section{Related Work}

The advancement of robotics is not limited to automation of hazardous, monotonous and repetitive tasks. In recent years, it has entered into the art-making industry to develop various forms of arts \citep{Jeon2017}. 

One of the robotic artist is humanoid robot painter \citep{ruchanurucks2007humanoid} that uses 2D vision technique for segmentation along with color perception model for accomplishing the drawing of portraits. In another work, \citep{jun2016humanoid} utilizes whole-body motion of a humanoid to paint an image on a wall using pointillism. More recently, a robotic system is shown to have a capability of producing a cartoonified version of human a portrait \citep{luo2018robot}, in which the robot paints like a human cartoonist by using non-photorealistic rendering (NPR) technique to generate hand-painted strokes. Another similar robotic artist, called \textit{Busker robot} \citep{scalera2019non}, used hatching and random splines as rendering techniques for aesthetic accomplishment.

To the best of authors' knowledge, the existing class of  work in this domain focus on the creation of art which a human can produce better. In this work, we use a robotic system to create a TSP-art which is not possible to draw by hand in feasible time.  In \citep{kaplan2005tsp}, the technique has been shown to convert a user-supplied image to construct a continuous line drawing using a TSP solver. In another work, on TSP-art \citep{bosch2008connecting} discusses how side constraints can be included to gain control of the interior and exterior of the tour. The existing works in the literature requires hit and trial for the generation of the final image and do not consider automation of the processes involved in generating the art. We propose to automate the process of art generation, delegate difficult craft-work to a robot and intake human aesthetic judgment to produce beautiful output.

\section{Chitrakar: System Overview}
\begin{figure}[h]
\centering
\includegraphics[width=0.5\linewidth]{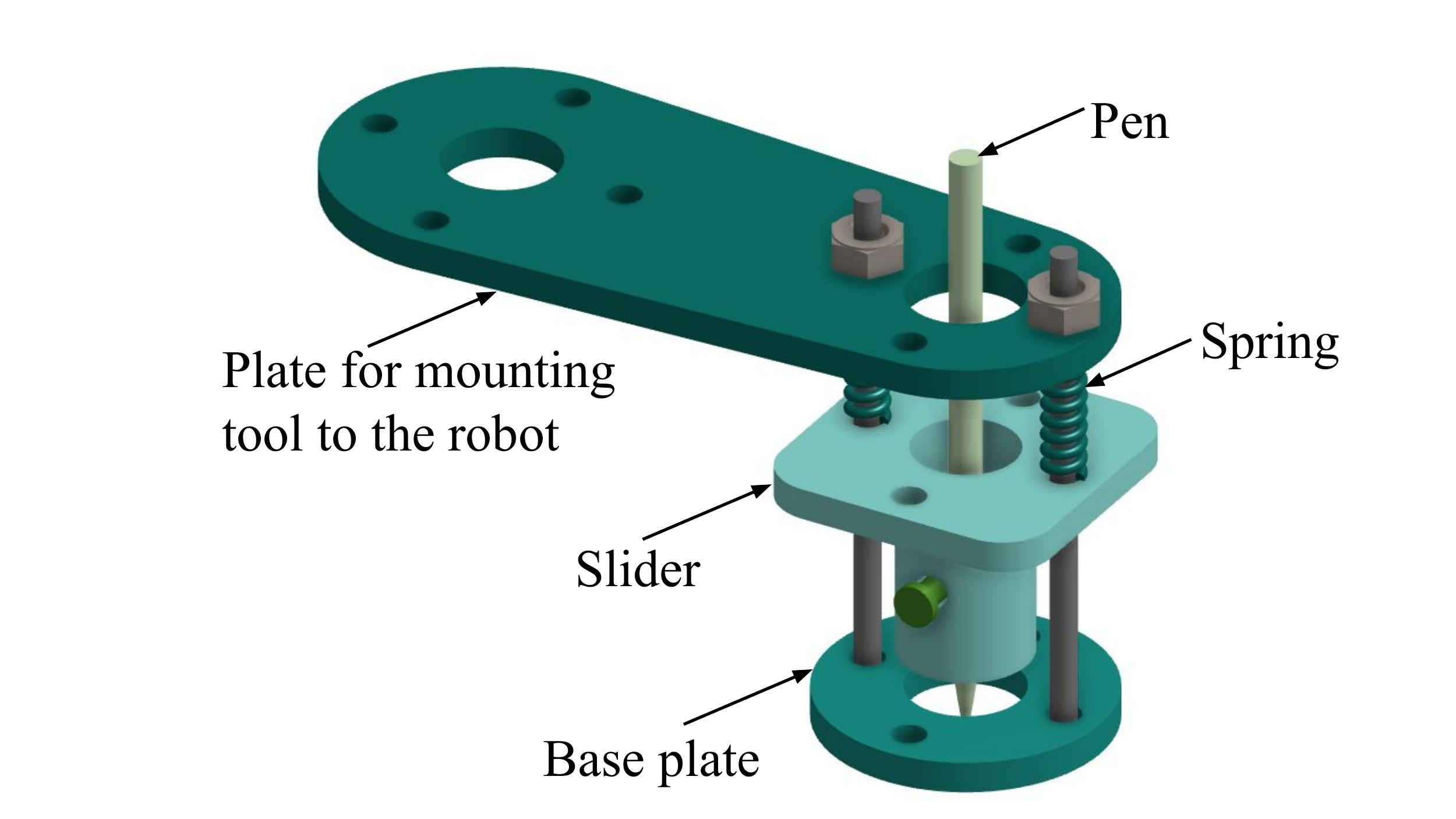}
\caption{\small 3D CAD design of the robot's end effector. It is designed to absorb movements along z-axis when a pen is moving on an uneven surface and ensures tear-free contact during the drawing process.}												\label{eef_design}
\end{figure}

 The system consists of a robotic assembly and a software pipeline shown in Fig. \ref{fig:flow_chart} and Fig. \ref{fig:real_system} respectively.
The hardware components consist of a six degrees-of-freedom UR5 robot arm, a drawing sheet of size 50$\times$100 cm and a customized end-effector for holding the pen as shown in Fig \ref{eef_design}. It moves on the xy-plane and allows small variation along z-axis due to the minor unevenness of the surface. An absence of spring mechanism will tear the paper, therefore to avoid this, the end-effector has an in-built slider.
A  high definition IP camera (Foscam FI9033) as shown in Fig. \ref{fig:real_system} is mounted on top of the arm to take a photograph of any human subject. 
The software consists of image processing and motion planning pipeline discussed in section \ref{sec:methodology}. The camera and robot are connected via a local network and can be controlled remotely using a computer. This computer has an Intel \textit{i7} processor running at 2.60GHz with 16GB RAM. 

\section{Methodology}
\label{sec:methodology}

The entire drawing process can be divided into four stages. Once the robot receives an input image it segments the person using Mask R-CNN. The segmented image is converted to a grayscale image, then enhanced using LoG and further converted to points which are connected using TSP-solver. The path generated by TSP-solver is made intersection free using a novel algorithm. This final output is drawn on a physical medium using the robotic assembly. The remaining part of this section will go through each step in detail.


\subsection{Instance segmentation}
\begin{figure}[b!]
\label{flow_chart}
\centering
\includegraphics[width=0.6\linewidth]{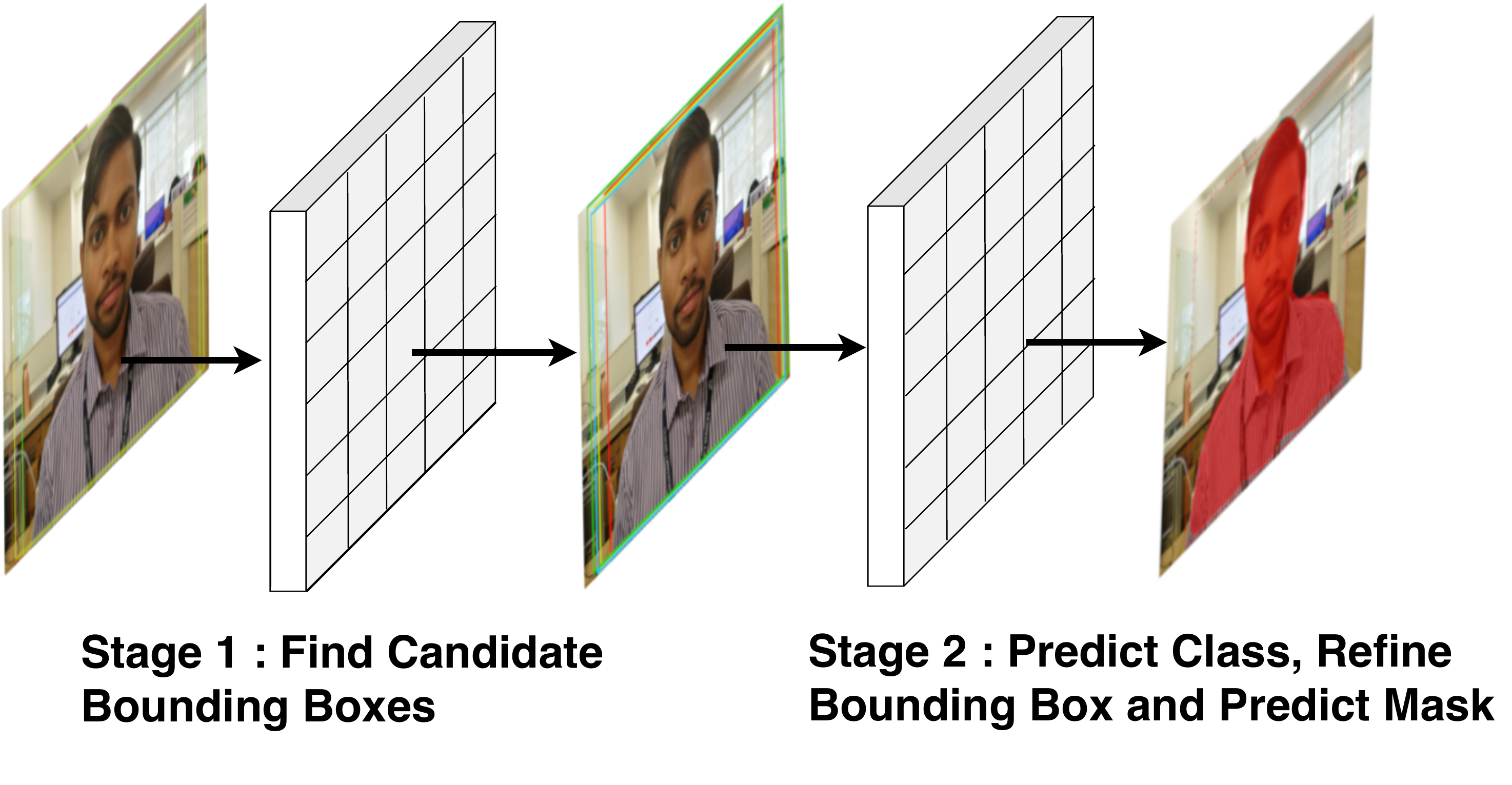}
\caption{\small Mask R-CNN Network Architecture.}
\end{figure}

						Instance segmentation is a technique which segments individual objects in an image by marking exact boundaries of the object. It is challenging because it requires correct detection of all the objects in an image and also computation of pixel-wise masks for every detected object in the image. We are using Mask R-CNN \citep{he2017mask} algorithm for instance segmentation and extraction of the mask of the desired object. 
Mask R-CNN is a two stage framework. In the first stage the network propose candidate object bounding boxes where there might be an object in the input image. In the second stage, it predicts the class of the object, refines the bounding box and predicts the mask in pixel level of the object. 
The resultant binary mask output generated from Mask R-CNN is used to subtract background from the image.
						
						
			
We have used Matterports code \citep{matterport_maskrcnn_2017} to extract mask of desired object from an input image. The Mask R-CNN model in Matterports code is trained on MSCOCO dataset. We have modified the code to predict only person's class in the supplied input image.

\subsection{Image processing}
For this application, we have created an image processing pipleine which will be followed by a probabilistic intensity based stippling. The steps are as follows -

\subsubsection{Image Enhancement} 
The segmented image is converted to grayscale image using Chromatic Adaptation Transform. The grayscale image is sharpened and enhanced to make the important features of the face prominent. The image enhancement is done using a Laplacian of Gaussian Filter. The Gaussian filter for 2D data is given in Eq. \ref{gaussian_equation}
\begin{equation}
	G(x,y)=\frac{1}{2\pi \sigma^2}e^{-\frac{x^2+y^2}{2\sigma^2}}
	\label{gaussian_equation}
\end{equation}

where, $\sigma=1$. As the $\sigma$ is increased the image has less noise and less detail. The output image of Gaussian filter is convolved with the Laplacian matrix of kernel having radius $r=1$. 
\begin{figure}[h!]
	\begin{center}
		\begin{tabular}{ccc}
			\includegraphics[width=0.2\linewidth]{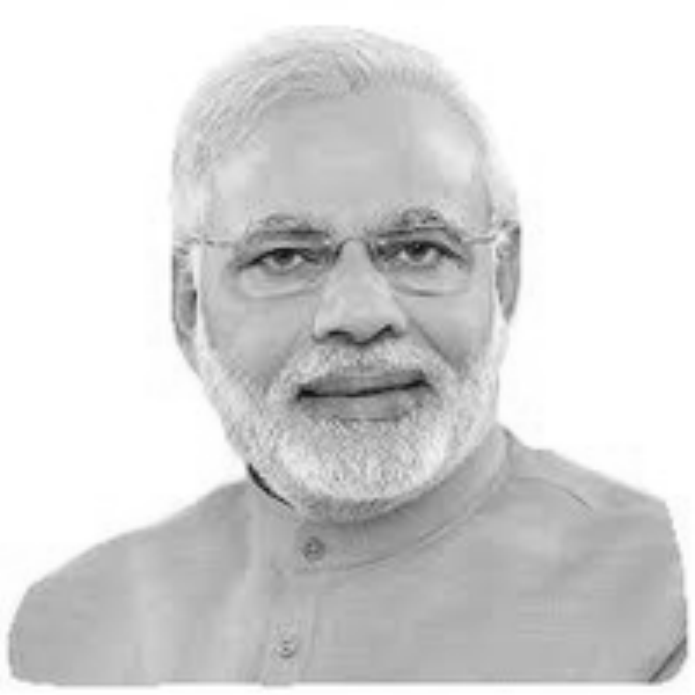}			\label{fig:modi_grayscale}&					
			\includegraphics[width=0.2\linewidth]{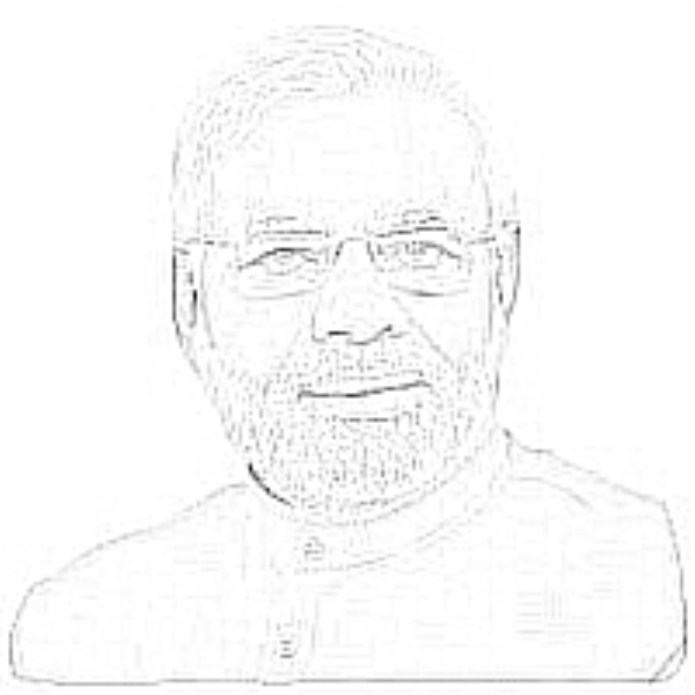} 		\label{fig:modi_laplace}&			    
			\includegraphics[width=0.2\linewidth]{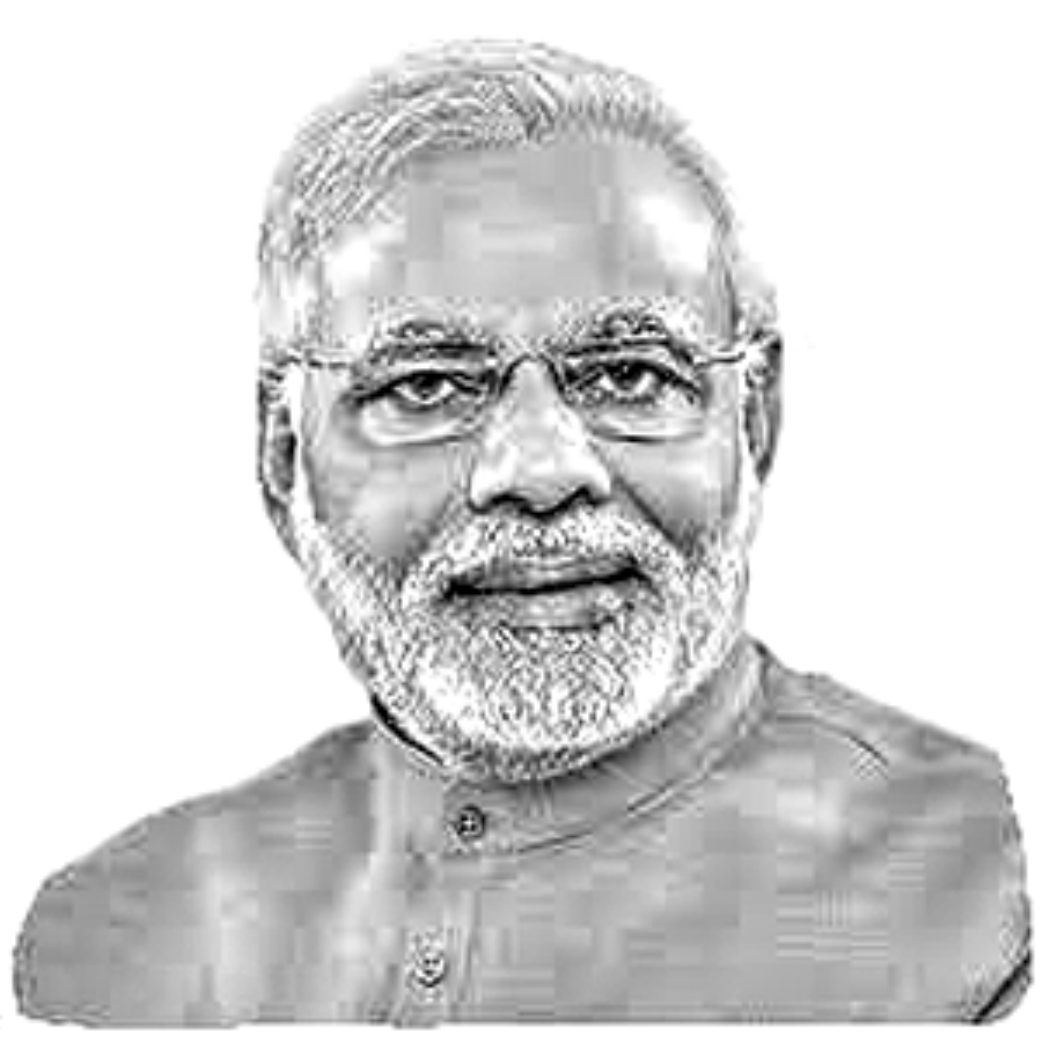} \label{fig:modi_en}
		\end{tabular}
	\end{center}
	\caption{Image Enhancement using Laplacian of Gaussian and image pixel multiplication. (a) shows the original grayscale image, (b) shows the Laplacian of Gaussian of the same image. Figure (c) shows the result of image multiplication of (a) and (b).}
	\label{fig:stippling}
\end{figure} 
\subsubsection{Intensity based probabilistic stippling} It is the process of creating a visual representation of an object using dots. In fine arts, the birth of this technique can be traced to Georges Seurat's technique of Pointillism in which he created painting by drawing dots of different colors on a canvas. It took him several years to complete each of his artworks. 

For TSP-art generation, it is essential to convert any given image to a set of points which can be used as the set of destinations for a travelling salesman. In stippling, sparse placement of dots to represent a lighter shade and high density of dots to represents darker shade is essential. 
The generated stippled images may be recognized as an art form in themselves but for creation of TSP-art the set of points of the stippled image is not good enough to create a recognizable representation of the person in the image. Other stippling techniques  gives more recognizable stippled images by varying dot size and use of colors. Such stippled images are not useful in this application because multiple colors and dot size is not a parameter that can be used by the robot while creating the art.
\begin{algorithm}[h]
	\caption{Intensity based probabilistic stippling}
	\label{alg:stippling}
	\begin{algorithmic}[1]
		\STATE Initialize processed image $I$ where $p_{i,j}$ corresponds to the intensity of the pixel at $(i,j)$
		
		\STATE $MaxIntensity = Max(p_{i,j} \forall p_{i,j} \in I)$
		
		\STATE $I$ = GrayScale($I$)
		\STATE $I$ = ImageEnhancement($I$)
		
		\FOR{ each pixel $p_{i,j}$ in image $I$}
		\IF{$p_{i,j}>MaxIntensity-Threshold$}
		\STATE $p_{i,j} = 1 $
		\ELSE 
		\STATE {$p_{i,j} = 0$}
		\ENDIF
		
		\ENDFOR
		
		\STATE Return $I$
	\end{algorithmic}
\end{algorithm}

\subsection{Jordan Curve Generation}

\subsubsection{TSP-solution} 		
After the image is stippled, the set of points is given as the input to travelling salesman solver. TSP is an NP-Hard problem and an optimal solution in polynomial time is not possible. Therefore, heuristics are used to obtain the solution of travelling salesman problem. Some heuristics like space-filling curves are very fast but cannot be used to create art and more optimal methods are time-consuming. 
\subsubsection{Intersection removal} We present a new method to create a Jordan Curve from existing TSP solution. The TSP solution is an ordered set of points which when joined in the given order creates a curve. The idea is to sequentially convert the TSP solution into an image and use that image to find intersections and remove them using 2-opt algorithm \citep{croes1958method}. The solution of TSP requires a weighted graph on which optimization can take place. It is a topological structure which is independent of the placement of vertices and edges. When TSP solvers are used to create a Jordan Curve, the challenge is to shift the points from a topological plane to a cartesian plane so that the intersection of edges can be taken into account. The switching of planes is a difficult feat and a workaround is to get a standard TSP solution and remove the intersection later. To check the intersection between lines, the exact coordinates of points are needed. One way to check the intersections is to start with the first line in the list and check intersection with every other line. In a set of $n$ lines, this will require $n!$ comparisons. These comparison increases exponentially for large values of $n$ and therefore makes this approach infeasible. 	

\begin{algorithm}[h]
	\begin{algorithmic}
		\STATE $T$ is the ordered set of points as a solution to TSP
		\STATE Initialize $d_1$ as maximum $x$ value in $T$ and $d_2$ as maximum $y$ value in $T$
		\STATE Initialize a null matrix $M$ of dimension $d_1$x$d_2$
		\STATE Initialize an empty set $Lines$
		
		\STATE  $intersection=True$ 
		\WHILE {$intersection==True$} 
		\STATE  $intersection=False$ 
		\FOR{$p_i$ and $p_{i+1}$ in $T$}
		\STATE $L_i$ = BresenhamLine($p_i$, $p_{i+1}$) 
		\FOR{$point$ in $L_i$}
		\IF{$M(point) == 1$}
		\STATE $L_j$=$2$-$Opt(L_i,Lines)$; 
		\STATE UpdateMatrix($M$,($L_i,L_j$))
		\STATE $intersection = True$
		\STATE \textbf{break}
		\ELSE 
		\STATE Add $L_i$ to $Lines$
		\STATE UpdateMatrix($	``M,(L_i)$)
		\ENDIF
		\ENDFOR
		\ENDFOR
		\ENDWHILE
		
	\end{algorithmic}
	\caption{Image based intersection removal}
	\label{alg:intersectionRemove}
\end{algorithm}

To reduce the number of intersection checks, we propose a novel method as given in Algorithm \ref{alg:intersectionRemove}. To check such intersections the process begins with an zero matrix of dimension $d_1 \times d_2$  and a TSP solution $T=\{(x_1,y_1), (x_2,y_2)... (x_n,y_n):x_i,y_i\in Z\}$ where $d_1$ is maximum $x$ value and $d_2$ is maximum $y$ value. Starting from the first point in $T$, a line is created between the current point $(x_i,y_i)$ and the next point $(x_{i+1},y_{i+1})$. The line is created using \textit{Bresenham's algorithm} \citep{bresenham1965algorithm}. For any two points $(x_i,y_i)$ and  $(x_{i+1},y_{i+1})$ in $T$, the Bresenham's algorithm gives a list of points $L_i$ through which the line will pass. For a set of $n$ points, $n-1$ lines will be created. The first and second line, $L_1$ and $L_2$, will never intersect but for lines from $L_3$ to $L_{n-1}$ intersections are possible. 

\begin{figure}[h]
	
	\begin{center}
		\begin{tabular}{ccc}
			\includegraphics[width=0.15\linewidth]{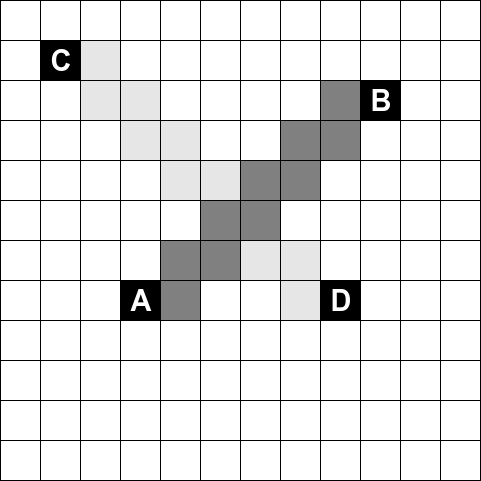}&
			\includegraphics[width=0.15\linewidth]{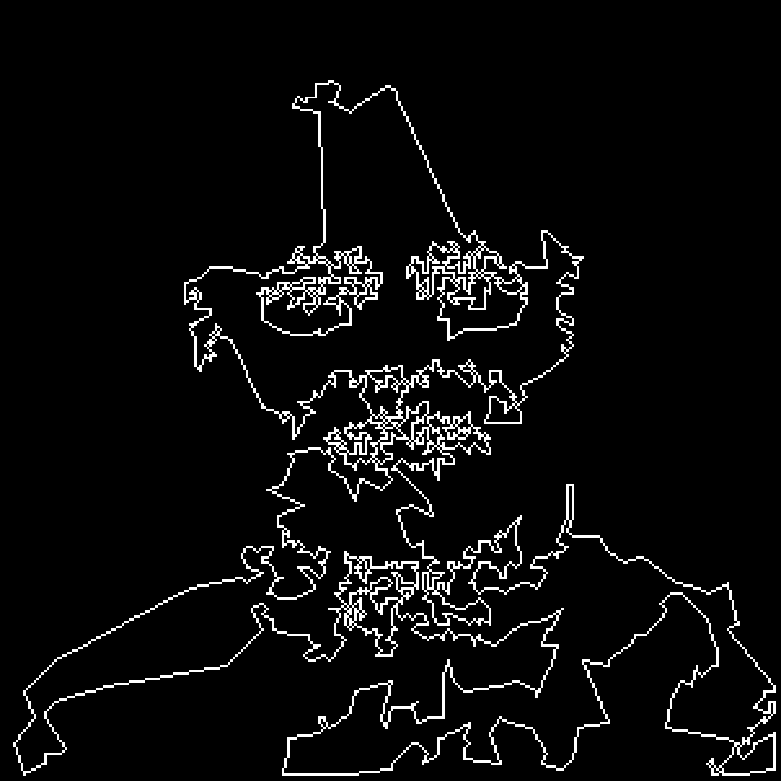}&			    
			\includegraphics[width=0.11\linewidth]{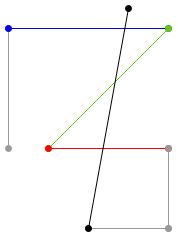}\\
			(a) & (b) & (c)\\		    
			
			\includegraphics[width=0.11\linewidth]{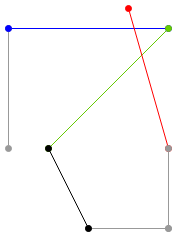}&
			\includegraphics[width=0.11\linewidth]{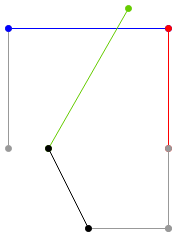}&			    
			\includegraphics[width=0.11\linewidth]{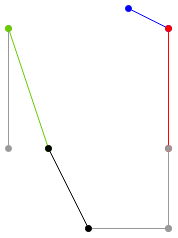}\\
			(d) & (e) & (f)
			
		\end{tabular}
	\end{center}
	\caption{The Fig. in (a) shows intersection of two lines in an image based data structure. In this data structure the dark pixels are stored as $1$ in an image (Matrix). Before drawing line `CD' we run Bresenham's algorithm to find any intersection with any pre-existing lines (`AB').
		The Fig. (b) to (f)  shows an instance of the process of intersection removal. The black line in Fig. (b) intersects with three lines. In Fig. (c) the intersection with of black and red line is removed by reconnecting the ends. The same is done for other intersections in Fig. (d) and Fig. (e). The final path is shown in Fig. (e) which is intersection free.}
	\label{fig:intersaction-removal}
\end{figure}

The intersection check starts by sequentially drawing the lines. Before drawing each line the value at the index of points in $L_i$ is checked. If it is one, the intersection is detected and the 2-opt algorithm is called to remove this intersection. The intersection removal process for one of the case of intersection is shown in Fig.  \ref{fig:intersaction-removal}. 
In a point storage system, to find the intersection of lines, any method will require a comparison of each line with every other line making it exponentially complex as the number of lines increases. With the visual representation of the image, to check an intersection only the values of the pixels on the line needs to be checked. If any of the pixels are already painted then an intersection is detected and 2-opt is used to remove that intersection.

\subsubsection{Human aesthetic judgement}
The process of creation of a Jordan Curve involves probabilistic selection of pixels because of which every generated Jordan Curve is different. The availability of computation power enables the generation of hundreds of different Jordan Curves in a few minutes. \textit{Chitrakar} exploits this computation power by generating $n$ Jordan Curves and put them on a screen for the user to select. Out of the $n$ Jordan Curves, the user can select the most aesthetically pleasing curve and let \textit{Chitrakar} draw it.

\subsection{Robot control}
For drawing the generated Jordan Curve on a paper, robotic arm motion is to be controlled according to the TSP solution. The input to motion planning module are the pixels coordinates of the portrait, which are first scaled and linearly transformed into the corresponding coordinates (in meters) with respect to the robot frame. The scaling operation can be performed as per the size and resolution requirement of the art. One must carefully perform the scaling operation as it would be difficult to acknowledge the art with inappropriate resolution. First, the path planning is done for smoothly moving the robotic arm through those points in a straight line. Then, the trajectories are planned by adopting trapezoidal velocity profiles for each motion. After the acceleration phase, the end-effector of the robot moves with constant velocity through the way-points and finally decelerates. 

The output of this module is the set of commands for the robot, written in the specific UR Script Programming Language in a specific script file, which includes all the instructions needed for drawing a portrait. 
Each command line in the script file defines the motion of end-effector of the robot and it consists of cartesian position coordinates in the configuration space, the orientation of the end-effector in axis-angle representation, the velocity, the acceleration, and the bend radius. The orientation of end-effector is kept fixed throughout the motion. The velocity and acceleration values should be carefully set up to avoid jerks in the motion of the robot. Finally, the drawing task is executed on the robot using a remote computer connected to the robot via Ethernet. It is worth noting that this module can successfully handle changes in the parameters such as resolution and stroke length without any change in the image processing output.

\section{Results and Discussion}

\begin{figure}[h]
	
	\begin{center}
		\begin{tabular}{ccc}
			\includegraphics[width=0.15\linewidth]{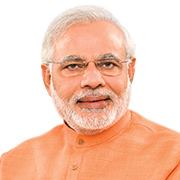}	&					
			\includegraphics[width=0.14\linewidth]{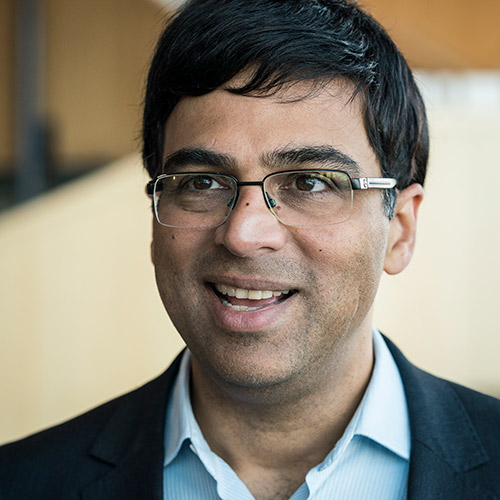}&			    
			\includegraphics[width=0.11\linewidth]{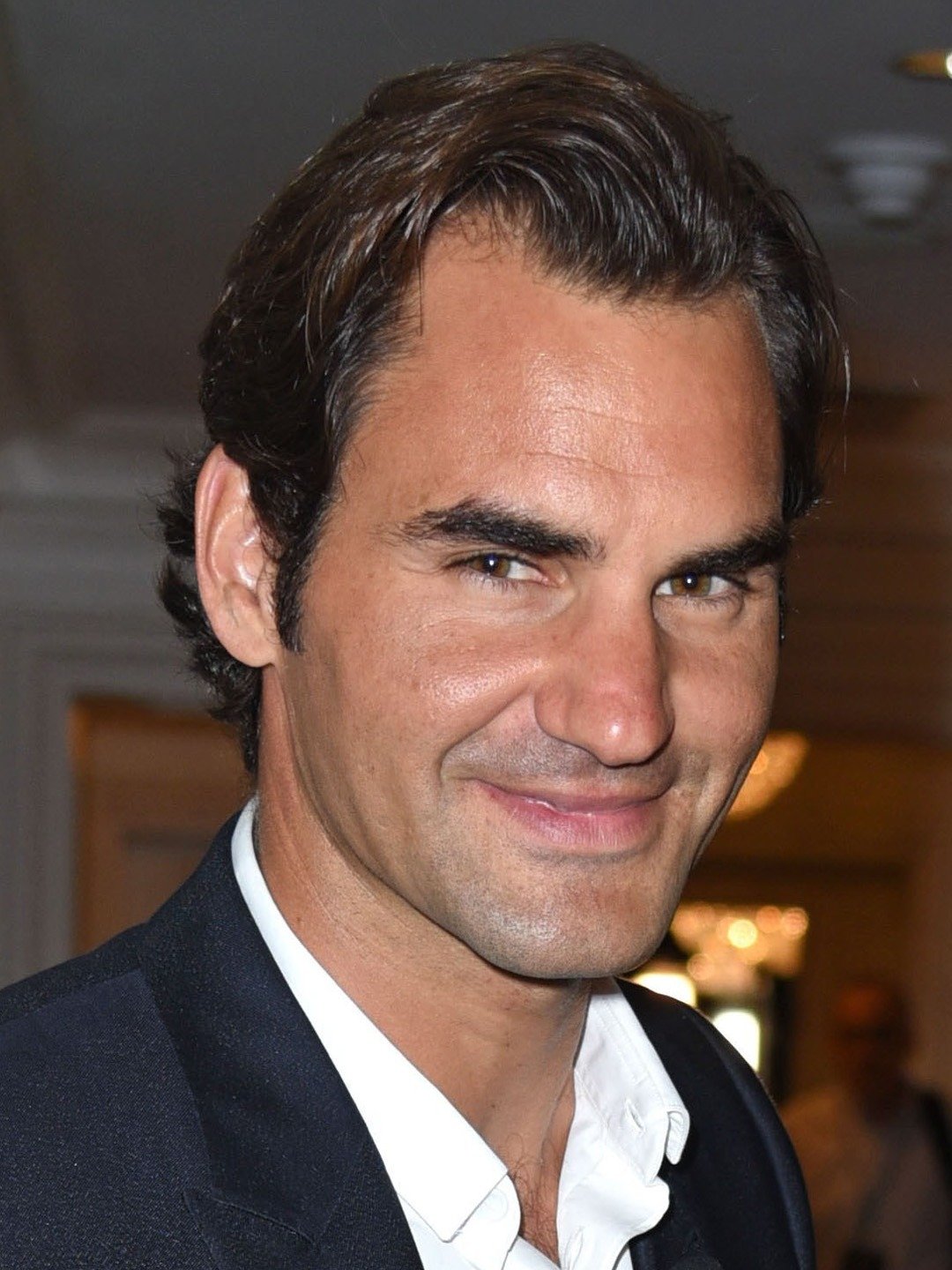}\\		    
			
			\includegraphics[width=0.15\linewidth, angle =270]{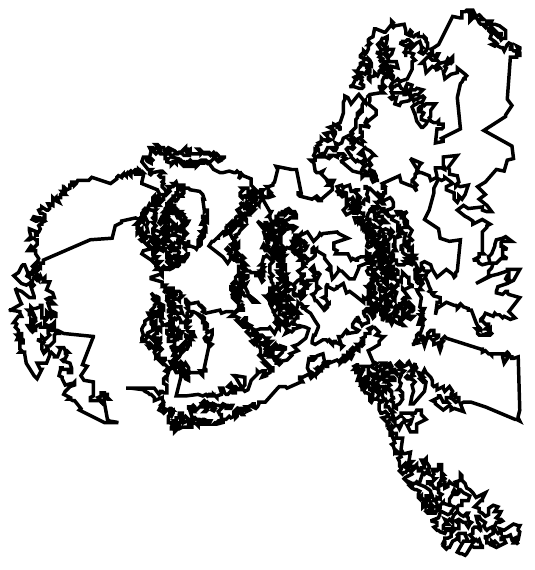}&
			\includegraphics[width=0.15\linewidth, angle =270]{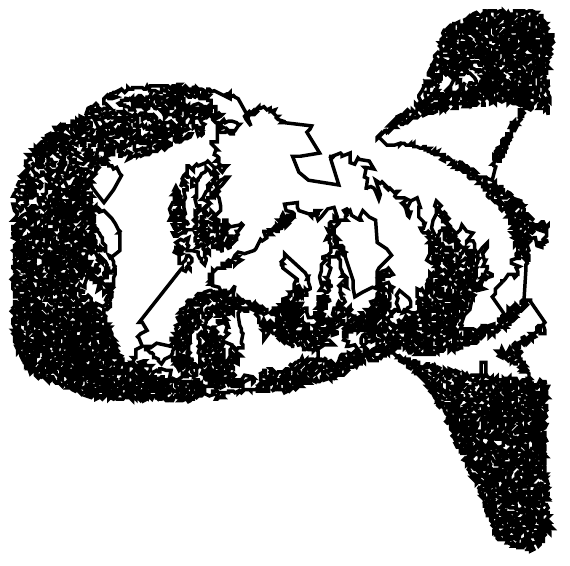}&			    
			\includegraphics[width=0.15\linewidth, angle =270]{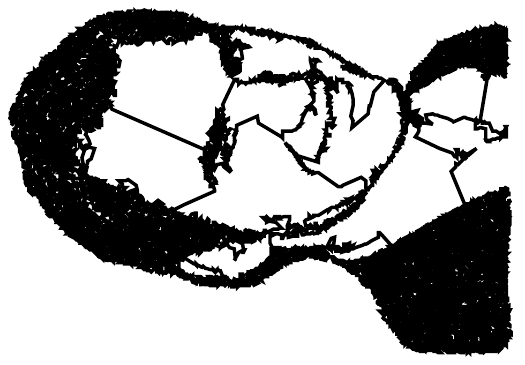}
		\end{tabular}
	\end{center}
	\caption{\small From left to right: Jordan Curve of Narendra Modi, Vishwanathan Anand and Roger Fedrer generated by using their publicly available photographs.}
	\label{fig:output_results}
\end{figure}

\begin{figure}[h]
	\begin{center}
		\begin{tabular}{ccc}
			\includegraphics[width=0.15\linewidth]{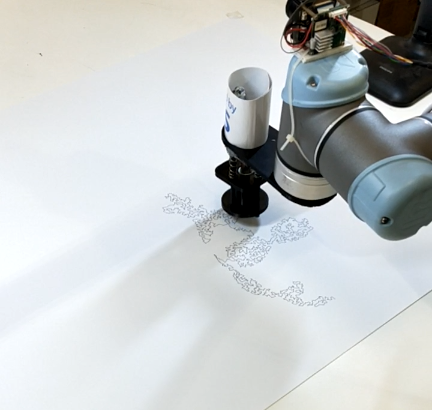}&			    
			\includegraphics[width=0.15\linewidth]{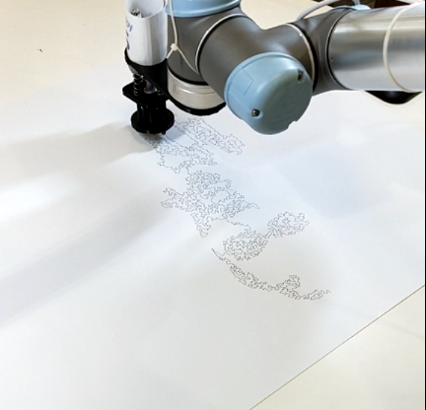} &	\includegraphics[width=0.15\linewidth]{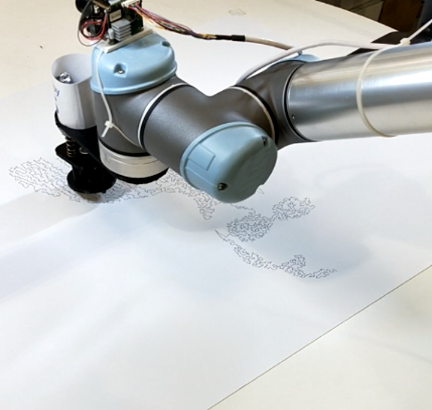} \\			    \includegraphics[width=0.15\linewidth]{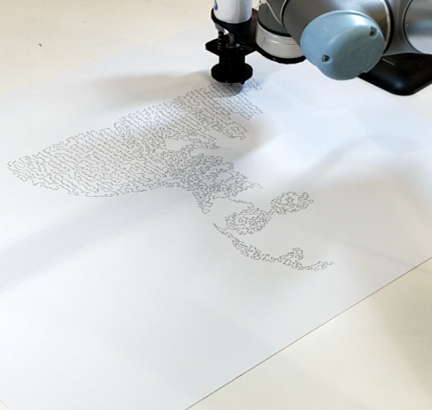} &
			\includegraphics[width=0.15\linewidth]{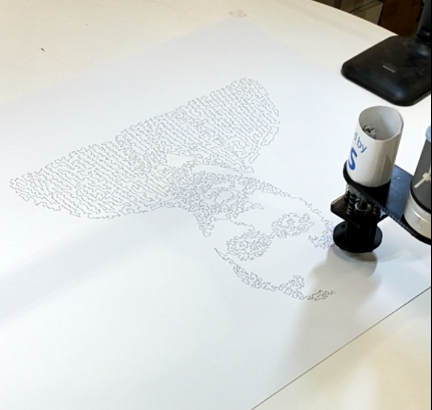} &
			\includegraphics[width=0.15\linewidth]{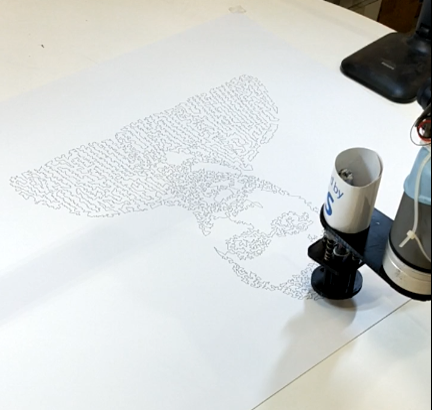}\\
		\end{tabular}
	\end{center}
	\caption{\small Drawing in process}
	\label{fig:draw_process}
\end{figure}

Several experiments have been conducted to test the efficacy of this system. Photographs of different people in different backgrounds with various lightening condition were fed into the system and the output is shown in Fig. \ref{fig:output_results}. The drawing process is shown in Fig. \ref{fig:draw_process}. The system is capable of converting any image to a Jordan Curve in real-time with a maximum time consumption within 1 minute.

\begin{figure}[h]
	\begin{center}
		\begin{tabular}{cc}
			\includegraphics[width=0.5\linewidth]{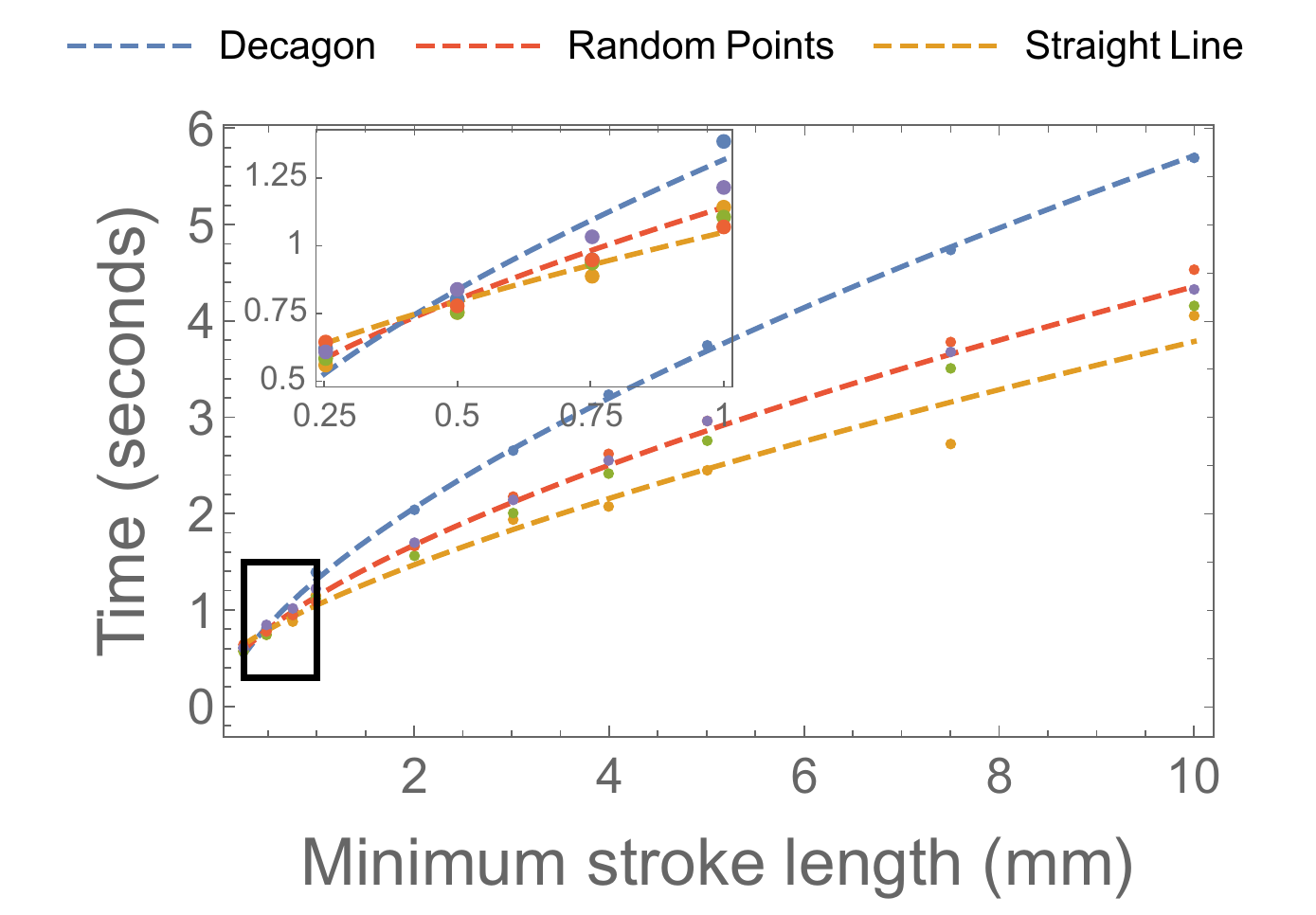} &
			\includegraphics[width=0.5\linewidth]{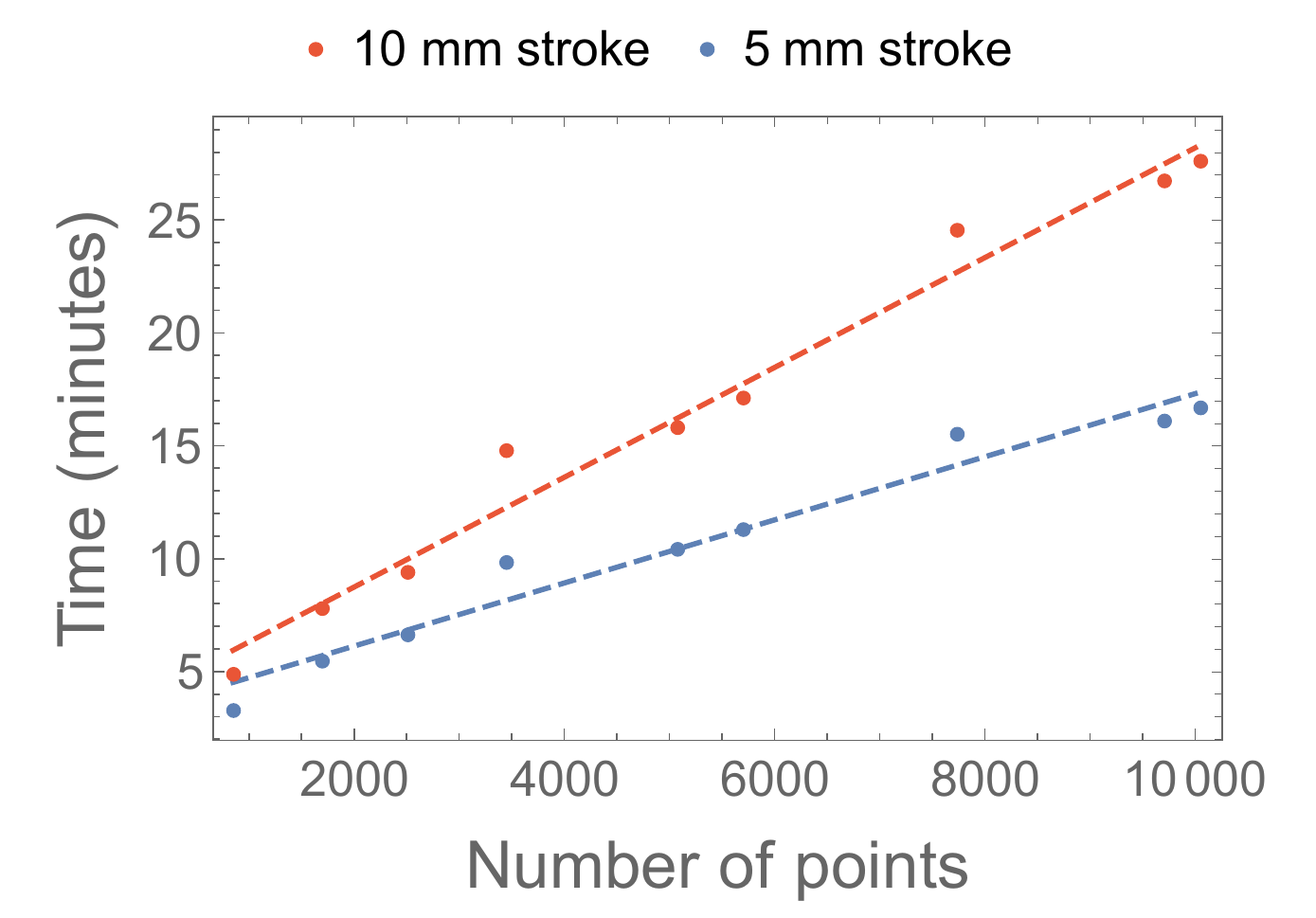}\\
			(a) & (b)
		\end{tabular}
	\end{center}
	\caption{(a) Time analysis for increase in stroke length  and (b) increase in number of points of Jordan curve.}
	\label{fig:timevsplots}
\end{figure} 
The time taken by the robot to draw is dependent on the number of points, resolution, and motion of the robot. To analyse the time taken to draw a portrait with different size and resolution, the system was tested by varying number of points in Jordan Curve and stroke length which is defined as the distance between two points. We considered three simple objects (straight line, decagon and randomly distributed points) for drawing to get an idea of upper and lower bounds of the time-taken to draw such objects.  First, we fix the number of points in each object to $10$ and vary stroke length. The result is shown in Fig. \ref{fig:timevsplots}(a). 
  The time taken by decagon is maximum because the path that needs to be travelled is long with maximum turns while for straight line it is minimum because the path is short and straight with no turns. 
  $$t_{Decagon}>t_{Random Points}>t_{Straight Line}$$
  A curve was fitted onto the data which is used to inform user about estimated time to draw portrait for different stroke length. The parameters of the best-fit curve are shown in Eq. 3.
\begin{equation}
\label{eq1}
f(x)=bx^a+c , where 
\end{equation}
\begin{equation}
      \left\{
  \begin{array}{lr}
    a=0.715,b=0.654,c=0.396 : Decagon\\
    a=0.626,b=1.36,c=-0.044 : Random\, Points\\
    a=0.643,b=0.945,c=0.195 : Straight\, Line
  \end{array}
\right.
\end{equation}

 
 Second, we fix the stroke length to 10mm and draw portrait with increasing the number of point ranging from 10 to 10000. The portraits were drawn on a $15\times15 cm$ sheet. The result is shown in Fig. \ref{fig:timevsplots}(b). 
  Straight line was fitted on the data with a slope of $m=0.00243$ and intercept $c=3.875$.
 A potrait can be created with any resolution depending upon the time availability of the person interacting with the system. The analysis showed that the maximum time for drawing a portrait with maximum stroke length of 10mm with a resolution as high as 10000 point will take about 27 minutes. 
\section{Conclusion and future work}
The fusion of multiple disciplines and techniques has resulted in a robotic system that can independently generate a Jordan Curve of any human face. The human intuitively selects one of the generated instances and then the robot draws it on a physical medium. The demonstration of this work confirms that the final output is recognizable as a human face of the subject and the result is aesthetically appealing. In a social context, this work makes human-robot relation more amicable and breaks the convention that robots can only be used for industrial application or for human replacement.

One of the limitations of this work is that the color conversion cannot be automated for an image which does not have a person in it. This is due to the use of Mask R-CNN for human detection which tunes this system specifically for human portrait generation. This limitation may be overcome by using more advanced image-processing techniques. 

As a future work, the system can be improved by developing a low-cost cartesian robot which can draw at a faster rate. 
The similar work can also be extended to three dimensions for sculpture making.

\bibliographystyle{unsrtnat}






\end{document}